%% file: main.tex
\documentclass[conference]{IEEEtran}
\IEEEoverridecommandlockouts
\usepackage{cite}
\usepackage{amsmath,amssymb,amsfonts}
\usepackage{algorithmic}
\usepackage{graphicx}
\usepackage{textcomp}
\usepackage{xcolor}
\usepackage{comment}
\usepackage[caption=false]{subfig}
\usepackage{color}
\usepackage[normalem]{ulem}
\usepackage{tcolorbox}
\usepackage{soul}

\newcommand{\shakshi}[1]{{\color{black}{\textbf{}}\color{black}{#1}}{\color{black}}}

\def\BibTeX{{\rm B\kern-.05em{\sc i\kern-.025em b}\kern-.08em
    T\kern-.1667em\lower.7ex\hbox{E}\kern-.125emX}}
\begin{document}

\title{Misinformation Concierge: A Proof-of-Concept\\ with \textcolor{black}{Curated Twitter Dataset on COVID-19 Vaccination}}

\author{\IEEEauthorblockN{Shakshi Sharma}
\IEEEauthorblockA{\textit{Institute of Computer Science,} \\
\textit{University of Tartu,}\\
Tartu, Estonia \\
shakshi.sharma268@gmail.com}
\and
\IEEEauthorblockN{Anwitaman Datta}
\IEEEauthorblockA{\textit{School of Computer Science and Engineering,} \\
\textit{Nanyang Technological University}\\
Singapore \\
anwitaman@ntu.edu.sg}
\and
\IEEEauthorblockN{Vigneshwaran Shankaran}
\IEEEauthorblockA{\textit{Department of Computer Science,} \\
\textit{University of Surrey,}\\
Guildford, UK \\
vigneshwaranpersonal@gmail.com}
\and
\IEEEauthorblockN{Rajesh Sharma}
\IEEEauthorblockA{\textit{Institute of Computer Science,} \\
\textit{University of Tartu,}\\
Tartu, Estonia \\
rajesh.sharma@ut.ee}

}

\maketitle

\begin{abstract}
We demonstrate the Misinformation Concierge\footnote{This is a preprint version of our CIKM paper}, a proof-of-concept that provides actionable intelligence on misinformation prevalent in social media. Specifically, it uses language processing and machine learning tools to identify subtopics of discourse and discern non/misleading posts; presents statistical reports for policy-makers to understand the big picture of prevalent misinformation in a timely manner; and recommends rebuttal messages for specific pieces of misinformation, identified from within the corpus of data - providing means to intervene and counter misinformation promptly. The Misinformation Concierge proof-of-concept using a curated dataset is accessible at: https://demo-frontend-uy34.onrender.com/
\end{abstract}

\begin{IEEEkeywords}
misinformation, social media, visual exploratory tool, decision support system, COVID-19 vaccines, Twitter
\end{IEEEkeywords}

\input{intro}

\input{related}

\input{arch}

\input{demo}

\input{conclusion}
\section*{Acknowledgment}

S. Sharma and R.Sharma's work has received funding from the EU H2020 program under the SoBigData++ project (grant agreement No. 871042), by the CHIST-ERA grant No. CHIST-ERA-19-XAI-010,  (ETAg grant No. SLTAT21096), and partially funded by HAMISON project. 

\bibliographystyle{IEEEtran}
\bibliography{main}

\end{document}

%% file: intro.tex
\section{Introduction}\label{sec:intro}

Misinformation is a major societal menace. Many vested interests are invested in manufacturing and propagating misinformation, including by engaging paid personnel to do so, besides co-opting `useful idiots', wreaking havoc on societies. 

Being able to monitor the kinds of misinformation gaining traction, and countering them using a dedicated team of personnel or (semi-)automated tools immediately rebutting rumors, particularly those with the propensity to persist, needs to be a part of the response and policing strategy against misinformation. The envisioned tool, \textbf{Misinformation Concierge}, is a step in that direction. Foremost, it captures the overall discourse pertaining to a broad subject of interest (for example, COVID-19 vaccines) from social media posts, e.g., Tweets. Employing language processing tools, it then identifies distinct (sub)topics and also discerns non/misleading posts, and visualizes the descriptive analysis results for users to better understand the big picture. Users could then drill down within specific topics (e.g., Pfizer vaccines) to investigate further aspects such as temporal behavior (e.g., which kind of topics are being discussed more or less over time?) as well as browse actual instances of misleading information pertaining to said topic. Finally, it allows one to zoom into individual misleading posts and explore post-specific information, e.g., which topic it belongs to, and identify other very similar misleading posts but also non-misleading posts. The latter could be used to refute the original offending post. Human users (or automated bots) could then repurpose these to promptly counter misleading posts. Such a platform would allow policymakers, as well as `first responders' in social media space, to both (i) identify and understand the prevalent misinformation in a timely and concise manner and also (ii) exploit ready to use responses derived automatically from the social media corpus itself.

We envision Misinformation Concierge as a general purpose platform to accommodate a variety of media and social media sources spanning a wider range of subjects to provide actionable intelligence. Still, in order to keep things tractable, the initial proof-of-concept is restricted to a curated Twitter dataset on COVID-19 vaccination \cite{sharma2021misleading}. 
The proof-of-concept using the curated dataset can be accessed at https://demo-frontend-uy34.onrender.com/.

%% file: related.tex
\section{Related Work}\label{sec:related}

Previous studies, e.g., \cite{balakrishnan2022infodemic,sharma2022facov}  identify many factors that contribute to the rapid digital dissemination of false news, including low factual understanding and an inability to recognize fake news. Thus, recommending correct information to the users is the first step to combat misinformation.
The detection of misinformation and interpretation of black-box models has been a major objective of existing misinformation studies \cite{mayank2022deap,sharma2021identifying,dhawan2022game}, with data analysis receiving little attention \cite{butt2022goes,jagtap2021misinformation}. Few works explore recommendation approaches to fake news \cite{wang2022veracity, you2019attributed}. However, in these cases, the recommendation is either in the form of the whole network \cite{kempe2003maximizing, nguyen2012containment} or based on fact-checking URLs exclusive to a small number of fact-checkers \cite{vo2018rise, you2019attributed}.
\shakshi{
The works that are specific to COVID-19 misinformation include releasing the new COVID-19 datasets \cite{hossain2020covidlies} or evaluating the deep learning models on publicly available datasets \cite{alhakami2022evaluating,kou2022hc}. Few studies also work on the impact of COVID-19 misinformation on mental health \cite{verma2022examining} and how to tackle effectively \cite{germani2022and}.}

The work closest to ours is the personalized recommendation \cite{wang2022veracity} which is based on user-level recommendations, as opposed to our tweet-centric recommendation approach. 
The limitation of a user-level approach is that if new events occur, it is difficult to collect the user's relevant reading history to personalize recommendations. A user might also want to explore new genres of news. Furthermore, user-level approaches risk creating a bubble or tunnel vision of their own. In these regards, a tweet-centric recommendations mechanism complements and can help. Specifically, suppose a user is interested in a certain post that happens to carry misinformation. In that case, our approach can extract the topics, sub-topics (in the form of entities), and the associated sentiment, and then it can recommend similar but factual non-misleading posts to the user, irrespective of the user's own biases or consumption history.
Thus, in this work, keeping in mind the issue of confirmation bias of social media, we focus on recommending users with correct information targeted on the same topic as a more practical and reasonable approach to combat spreading misinformation on social media.

%% file: arch.tex
\begin{figure*}[t]
\includegraphics[width=0.9\textwidth, height=5.4cm, trim={0cm 4cm 0cm 3cm},clip]{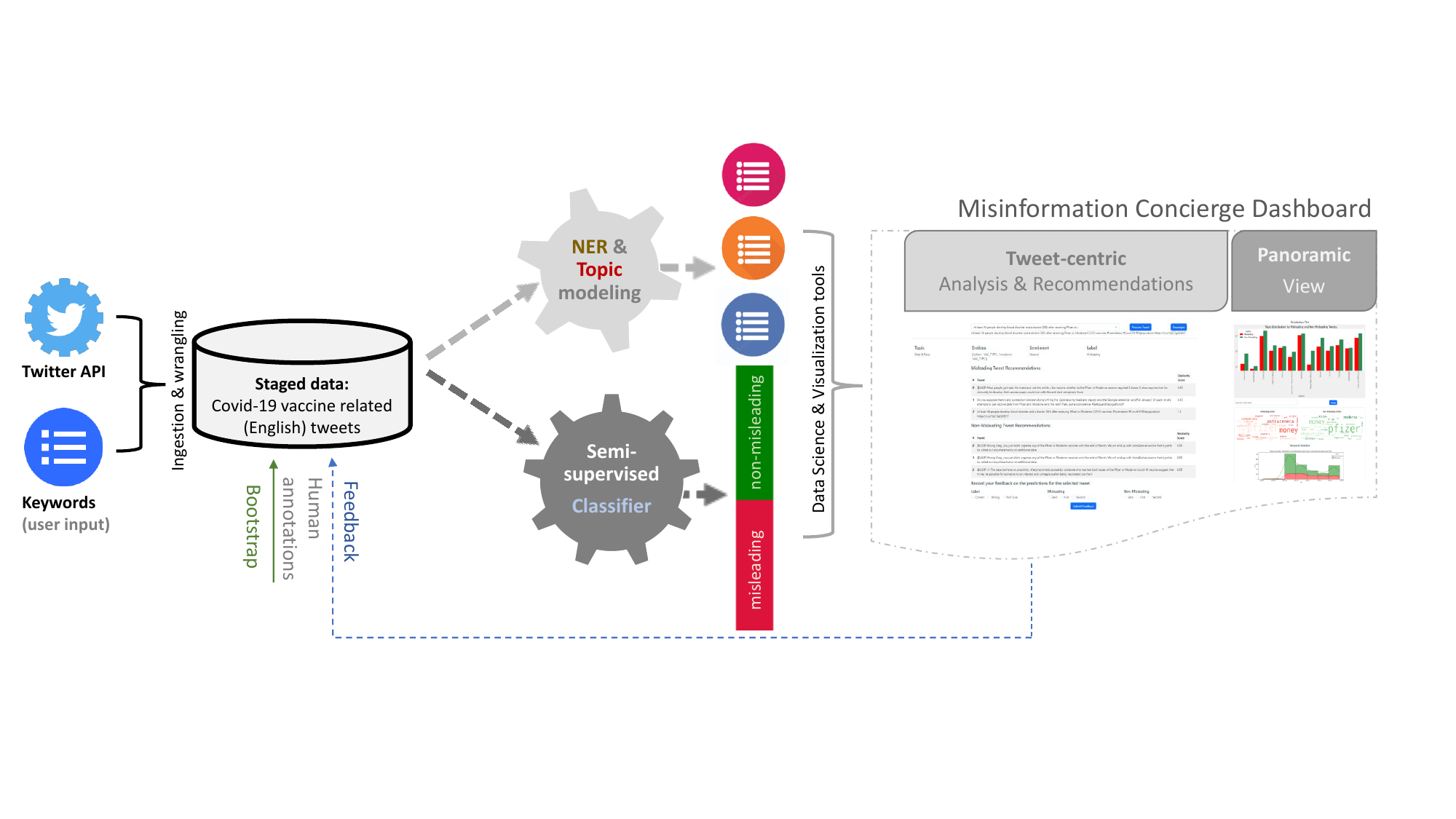}
\centering
\caption{A high-level view of the Misinformation Concierge architecture.}
\label{fig:arch}
\end{figure*}

\section{System Architecture}\label{sec:arch}

Misinformation Concierge has a modular architecture (Figure \ref{fig:arch}). The individual tasks, e.g., data staging, classification, topic extraction, etc., can be accomplished using a different mechanism than what we currently use so as to organically upgrade the system with better tools for individual tasks. e.g., with better quality of results. The emphasis of this demo is as such on the overall framework, rather than maturity and quality of the solutions for the individual tasks. Next, we explain how these tasks are accomplished currently. 

\subsection{Data acquisition, staging \& classification}
We use a publicly accessible dataset \cite{sharma2021misleading} for this work, which contains 114,635 COVID-19 vaccination related tweets for the seven months from September 2020 to March 2021 labeled as misleading or non-misleading using ML techniques trained with originally 1,500 manually annotated data. 

In general, to deploy Misinformation Concierge on a new subject, a user would (1) need to provide relevant keywords to be used for acquiring Tweets through Twitter API, which is then (2) cleaned mostly through automated tools, e.g., removing non-English tweets; but also identifying tweets which match a single keyword from step-1, which might need a manual inspection to determine relevance to the subject, e.g., for the chosen case study, some vaccine related tweets were not relevant to COVID-19 and had to be eliminated. Finally, the collected dataset has to be (3) classified as non/misleading using a semi-supervised classifier \cite{sharma2021misleading}, which in turn needs a small subset of manually labeled data. 

Subsequently, as the system is used, end-users could provide feedback on the quality of recommendation in a low-effort manner, to be used to transparently expand the manually annotated data and regularly rerun and refine the ML models in the background. 

\subsection{Recommendation of Non-Misleading Tweets}

To recommend related non-misleading tweets to counter misleading tweets, we use three Natural Language Processing (NLP) techniques: LDA topic modeling \cite{sharma2021misleading} and Named Entity Recognition\footnote{https://spacy.io/api/entityrecognizer} (NER) for extracting general and specific topics, accompanied by matching sentiments\cite{sharma2021misleading} of the tweet.

\subsubsection{General (Broad) Topic extraction}

It is crucial first to understand the broad topic of a given tweet. 
In this work, similar to \cite{sharma2021misleading}, we utilize LDA topic modeling to assign topics to each tweet. The names of the topics are manually ascribed following automatic detection. 
Specifically, if a particular topic is present in the tweet, we label it as one of the topics. 
This approach gives us 12 topics in total, and 4,063 tweets are left without labels (Unknown label). To further improve our process for the Unknown label, we used synonyms\footnote{https://www.nltk.org/howto/wordnet.html}
for each topic and checked if the synonyms of each topic were present in the tweets that contained an Unknown label. 817 tweets could thus be labeled as one of the 12 topics, leaving 3,426 tweets with an Unknown label, as shown in Table
\ref{tbl:topics}.

\begin{table}[!htbp]
\footnotesize
\centering
\caption{Absolute number (and \%) of Tweets per topic. }
\label{tbl:topics}
\begin{tabular}{|l|l|}
\hline
\textbf{Topics}      & \textbf{Tweets: number (\%)} \\ \hline
Choices              & 33,150 (28.9\%) \\ \hline
Politics             & 25,276 (22\%)   \\ \hline
Vaccine Efficacy     & 21,936 (19.1\%) \\ \hline
Shots                & 9,568 (8.34\%)  \\ \hline
Trump                & 8,432 (7.35\%)  \\ \hline
Data \& Facts        & 3,601 (3.14\%)  \\ \hline
Unknown              & 3,426 (2.98\%)  \\ \hline
Trials               & 3,217 (2.8\%)   \\ \hline
Myths                & 2,376 (2\%)     \\ \hline
Operation Warp Speed & 1,369 (1.19\%)  \\ \hline
Real Side-Effects    & 1,216 (1.06\%)  \\ \hline
Approval             & 883 (0.77\%)    \\ \hline
Availability         & 185 (0.16\%)    \\ \hline
\end{tabular}%
\end{table}

\subsubsection{Specific Topic Extraction}
Next, we dig deeper into the tweets to find named entities used in each tweet.
We explored various NER (Named Entity Recognition) models, for example, the latest pretrained roberta-base NER  model, en\_core\_web\_trf (which identified entities in 57\% of our data). The NER model  en\_core\_web\_sm gave us the best performance containing 88\% data with at least one entity.
We observed two problems with the use of such pre-existing models. First, there are words specific to COVID-19, such as Pfizer, Shots, and Johnson \& Johnson, that were being wrongly labeled as PERSON, GPE (Geopolitical entity). Second, there were still 12\% of rows that contained zero (or null) entities. 

Consequently, we fine-tuned the NER model on a random subset of manually labeled 100 instances from the dataset that contained non-null entities. We added a new entity type called VAC\_TYPE for the words that contain 
vaccine names using a list of manually labeled VAC\_TYPE entities: [pfizer, astrazeneca, mrna, astrazenca, jnj, oxford, sputnik, modern, variants, \#pfizer, booster, \#astrazeneca, biontech, Covidshield]. To make the model robust and based on our previous observations, we also included spelling errors in the entities, such as `modern' instead of moderna, in the list. Additionally, we did not remove the hashtags from the dataset as the hashtags play a major role in identifying misinformation these days.
We trained the augmented model for 30 epochs with a dropout rate of 0.3 utilizing five-fold cross-validation. 
Next, we compare the performance of the augmented (fine-tuned) model with the unaugmented model on the same subset of manually labeled 100 examples shown in Table \ref{tbl:comparison}. We observe that the augmented model performs better in all the metrics with a good margin. Finally, we labeled all the tweets in the dataset with the augmented model.
We obtained the list of VAC\_TYPE entities that are labeled by the augmented model, which included new entries beside the ones we had manually provided during training: [phizer, myrna, zenca, novavax, johnsonandjohnson, johnson, mirna]. 

\begin{table}[!htbp]
\centering
\caption{Comparison of the Spacy NER model with and without fine-tuning the model on the manually labeled data.}
\label{tbl:comparison}
\resizebox{0.8\columnwidth}{!}{%
\begin{tabular}{|l|l|l|l|l|}
\hline
\textbf{Spacy model} & \textbf{Accuracy} & \textbf{Precision} & \textbf{Recall} & \textbf{F1 Score} \\ \hline
w/o training         & 0.27              & 0.25               & 0.26            & 0.25              \\ \hline
w/ training          & 0.89              & 0.90               & 0.88            & 0.87              \\ \hline
\end{tabular}%
}
\end{table}

We notice that 112,994 (98.5\%) rows have non-identical entities with the augmented model (up from 88\% with the unaugmented one), indicating a profound improvement in the detection of entities and entity type. Among the 36,829 previously mislabeled entities, 35,480 (96.3\%) and 1,349 (3.7\%) entities represent VAC\_TYPE and non-VAC\_TYPE entities. Table \ref{tbl:mislabeled} represents the top entities divided into  VAC\_TYPE and non  VAC\_TYPE entities that are mislabeled by the unaugmented model. For example, as shown in Table \ref{tbl:mislabeled}, the entity Pfizer has been incorrectly labeled (in the given context) as ORG by the unaugmented model and correctly labeled as VAC\_TYPE by the augmented model. VAC\_TYPE entities detected by the model are in such a big number since most of the dataset contains vaccine names in their tweets.


In very few instances, some entities were labeled correctly by the unaugmented model but incorrectly by the augmented model. Particularly, a few geographic places were mislabeled as VAC\_TYPE by the augmented model but correctly as GPE by the unaugmented model. Additionally, there have been cases where both models fail to identify the entity accurately, e.g., the un/augmented models have incorrectly labeled Ohio as MONEY and VAC\_TYPE, respectively.

\begin{table}[!htbp]
\footnotesize
\centering
\caption{Top entities mislabeled by the unaugmented NER model. The left and right sides represent vaccine names and non-vaccine names entities.
}
\label{tbl:mislabeled}
\resizebox{\columnwidth}{!}{%
\begin{tabular}{|ccc|ccc|}
\hline
\multicolumn{3}{|c|}{\textbf{VAC\_TYPE}}                                                                & \multicolumn{3}{c|}{\textbf{Others}}                                                               \\ \hline
\multicolumn{1}{|c|}{\textbf{Entity}}     & \multicolumn{1}{c|}{\textbf{Mislabeled}} & \textbf{Correct} & \multicolumn{1}{c|}{\textbf{Entity}} & \multicolumn{1}{c|}{\textbf{Mislabeled}} & \textbf{Correct} \\ \hline
\multicolumn{1}{|c|}{pfizer}              & \multicolumn{1}{c|}{ORG}                 & VAC\_TYPE        & \multicolumn{1}{c|}{millions}        & \multicolumn{1}{c|}{CARDINAL}            & MONEY            \\ \hline
\multicolumn{1}{|c|}{moderna}             & \multicolumn{1}{c|}{GPE}                 & VAC\_TYPE        & \multicolumn{1}{c|}{billions}        & \multicolumn{1}{c|}{CARDINAL}            & MONEY            \\ \hline
\multicolumn{1}{|c|}{astrazeneca}         & \multicolumn{1}{c|}{ORG}                 & VAC\_TYPE        & \multicolumn{1}{c|}{trump}           & \multicolumn{1}{c|}{ORG}                 & PERSON           \\ \hline
\multicolumn{1}{|c|}{johnson and johnson} & \multicolumn{1}{c|}{PERSON}              & VAC\_TYPE        & \multicolumn{1}{c|}{biden}           & \multicolumn{1}{c|}{ORG}                 & PERSON           \\ \hline
\multicolumn{1}{|c|}{novavax}             & \multicolumn{1}{c|}{ORG}                 & VAC\_TYPE        & \multicolumn{1}{c|}{lock down}       & \multicolumn{1}{c|}{NORP}                & EVENT            \\ \hline
\end{tabular}%
}
\end{table}



\subsubsection{Sentiments Extraction}
Users suffering 
from confirmation bias tend to prefer to read posts on a certain topic that reflect their sentiments. 
We employ VADER API\cite{sharma2021misleading} to detect the sentiment of the tweet as positive, negative, or neutral.

\subsubsection{Recommend Similar (Non-)Misleading Tweets}
To determine the similarity between a given misleading tweet and other (non-)misleading tweets, we tested two approaches: direct string matching (using methods like Hamming distance) and vector dimension matching (such as Word2Vec) to match the misleading tweets with equivalent (non-)misleading tweets. The optimal similarity matching for our data was obtained with vector dimension matching. After experimenting with several combinations, we utilized GloVe embedding\cite{pennington2014glove} with cosine similarity in particular.

For a given Misleading tweet, we use three criteria, i.e., the general topic, entities extracted by the augmented NER model, and the sentiment. Next, we match these three criteria over the corpus of the (non-)Misleading tweets. Once such (non-)Misleading tweets have been retrieved, the top-K similar (non-)Misleading tweets are identified using the GloVe embedding and cosine similarity. 

%% file: demo.tex
\section{Demonstration}\label{sec:demo}
\begin{figure*}[t!]
\begin{tabular}{|c|c|}
\hline
Panoramic view: A view of the big picture & Tweet-centric analysis \& recommendation \\\hline
\subfloat{\includegraphics[width=0.99\columnwidth, height = 6.5cm, trim={0cm 0cm 0cm 0cm},clip]{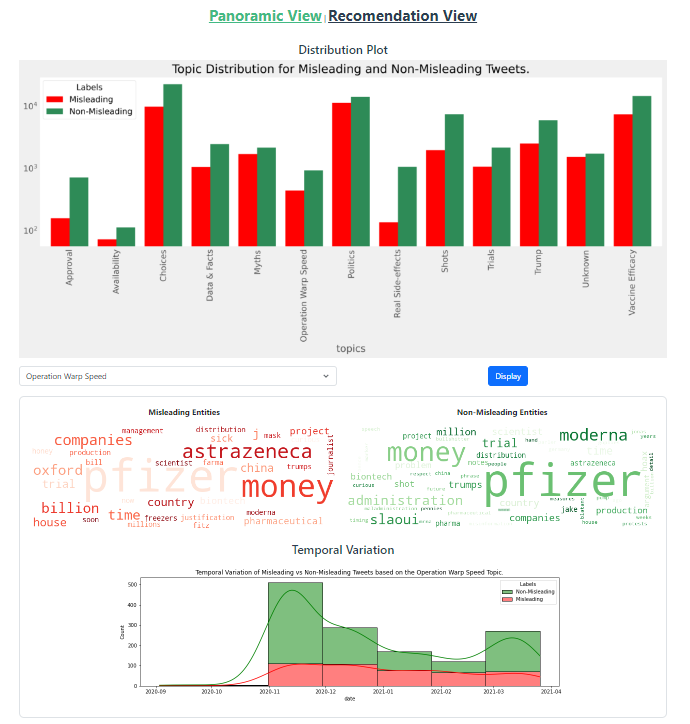}} &
\subfloat{\includegraphics[width=0.99\columnwidth, height = 6.5cm]{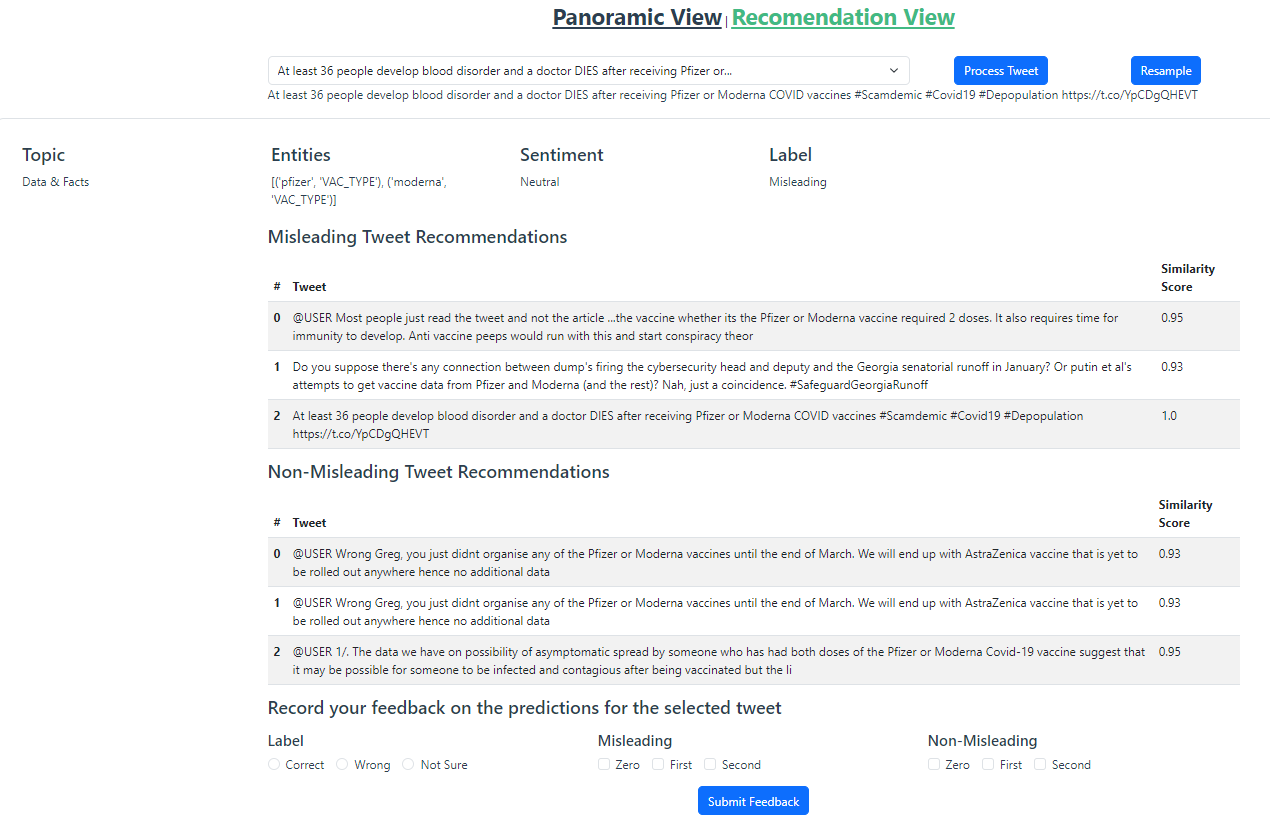}}\\\hline
\end{tabular}
\caption{Recommendation Dashboard. The left and right sides represent panoramic and tweet-centric views. The NM and M denote the Non-Misleading and Misleading tweets, respectively.}
\label{fig:demo}
\end{figure*}

For the demonstration, we use a dataset preprocessed as discussed above, which includes columns for sentiments, topics, entities, labels, and tweets. During the live demo, we will explain these steps while showcasing the Misinformation Concierge functionalities comprising broadly two parts:  A Panoramic View capturing the big picture and a Tweet-centric Analysis \& Recommendation View.

\textbf{The Panoramic View} provides an overview of the dataset.
In particular, displaying the distribution of the topics for (non-)misleading labels in the dataset serves as the first step.
Based on the topic selected by the user, word clouds of entities across both labels are displayed, along with temporal characteristics of how and when the topic has gained traction. 

\textbf{The Tweet-centric Analysis \& Recommendation View} analyses individual tweets, identifying its topic, involved entities, sentiments, identifies other similar misleading tweets, and recommends non-misleading tweets that can be used to counter the chosen tweet. The number of such recommendations is a user chosen parameter (with a default set to three). 
Finding other similar misleading tweets could be used to identify homophilic users and echo chambers, while the non-misleading tweets can be reused for rebuttal. 

The underlying algorithms are inherently imperfect, furthermore, they have been trained with very small samples of human annotated data to bootstrap the system. To ameliorate the latter issue, the dashboard captures feedback from the end-user on the correctness of labels assigned to tweets. This is used as feedback to generate more human annotations organically over time without the onerous burden of labeling data explicitly. The underlying models can thus be retrained and refined over time as the system is used (this is implemented but not tested, since our system currently does not have a user base), and augmented with extrinsic resources such as large language models to improve the quality of recommendations and incorporate live data in the future. 

\shakshi{
\textbf{Challenges:}
The deployment of the classification model might run into a notable challenge known as the "cold start" problem, which could hamper its capacity to effectively categorize non-COVID-19 topics. However, by employing a fine-tuning approach to the classification model using even a small dataset particularly labeled for non-COVID-19 topics, this issue can be efficiently overcome. This approach makes it possible to greatly improve the model's performance across diverse datasets. In addition, user feedback about misclassified examples would help in model refinement.
The second challenge is that the effectiveness of the demo's underlying components, such as the LDA, NER, and semi-supervised learning modules, is closely related to the utility of the demo. In fact, the demo's current version makes use of conventional methods like VADER and a model that was trained on annotated examples. We purposefully chose to demonstrate functionality above seeking state-of-the-art (SOTA) performance because our goal was to offer a working model that could be utilized as a starting point. With this strategy, we are able to establish a foundation for future development while also providing useful insights into the possibilities of more advanced methods in subsequent iterations.
The third challenge involves the recent changes in Twitter API, we are committed to actively monitoring and adapting the tool to align with the most recent developments and ensure its continued relevance and utility. 
}

%% file: conclusion.tex

%% file: main.bbl
\begin{thebibliography}{10}
\providecommand{\url}[1]{#1}
\csname url@samestyle\endcsname
\providecommand{\newblock}{\relax}
\providecommand{\bibinfo}[2]{#2}
\providecommand{\BIBentrySTDinterwordspacing}{\spaceskip=0pt\relax}
\providecommand{\BIBentryALTinterwordstretchfactor}{4}
\providecommand{\BIBentryALTinterwordspacing}{\spaceskip=\fontdimen2\font plus
\BIBentryALTinterwordstretchfactor\fontdimen3\font minus
  \fontdimen4\font\relax}
\providecommand{\BIBforeignlanguage}[2]{{%
\expandafter\ifx\csname l@#1\endcsname\relax
\typeout{** WARNING: IEEEtran.bst: No hyphenation pattern has been}%
\typeout{** loaded for the language `#1'. Using the pattern for}%
\typeout{** the default language instead.}%
\else
\language=\csname l@#1\endcsname
\fi
#2}}
\providecommand{\BIBdecl}{\relax}
\BIBdecl

\bibitem{sharma2021misleading}
S.~Sharma, R.~Sharma, and A.~Datta, ``(mis)leading the covid-19 vaccination
  discourse on twitter: An exploratory study of infodemic around the
  pandemic,'' \emph{IEEE Transactions on Computational Social Systems}, 2022.

\bibitem{balakrishnan2022infodemic}
V.~Balakrishnan, N.~Zhen, S.~Chong, G.~Han, and T.~Lee, ``Infodemic and fake
  news--a comprehensive overview of its global magnitude during the covid-19
  pandemic in 2021: A scoping review,'' \emph{International Journal of Disaster
  Risk Reduction}, p. 103144, 2022.

\bibitem{sharma2022facov}
S.~Sharma, E.~Agrawal, R.~Sharma, and A.~Datta, ``Facov: Covid-19 viral news
  and rumors fact-check articles dataset,'' in \emph{Proceedings of the
  International AAAI Conference on Web and Social Media}, vol.~16, 2022, pp.
  1312--1321.

\bibitem{mayank2022deap}
M.~Mayank, S.~Sharma, and R.~Sharma, ``Deap-faked: Knowledge graph based
  approach for fake news detection,'' in \emph{2022 IEEE/ACM International
  Conference on Advances in Social Networks Analysis and Mining
  (ASONAM)}.\hskip 1em plus 0.5em minus 0.4em\relax IEEE, 2022, pp. 47--51.

\bibitem{sharma2021identifying}
S.~Sharma and R.~Sharma, ``Identifying possible rumor spreaders on twitter: A
  weak supervised learning approach,'' in \emph{2021 International Joint
  Conference on Neural Networks (IJCNN)}.\hskip 1em plus 0.5em minus
  0.4em\relax IEEE, 2021, pp. 1--8.

\bibitem{dhawan2022game}
M.~Dhawan, S.~Sharma, A.~Kadam, R.~Sharma, and P.~Kumaraguru, ``Game-on: Graph
  attention network based multimodal fusion for fake news detection,''
  \emph{arXiv preprint arXiv:2202.12478}, 2022.

\bibitem{butt2022goes}
S.~Butt, S.~Sharma, R.~Sharma, G.~Sidorov, and A.~Gelbukh, ``What goes on
  inside rumour and non-rumour tweets and their reactions: A psycholinguistic
  analyses,'' \emph{Computers in Human Behavior}, vol. 135, p. 107345, 2022.

\bibitem{jagtap2021misinformation}
R.~Jagtap, A.~Kumar, R.~Goel, S.~Sharma, R.~Sharma, and C.~P. George,
  ``Misinformation detection on youtube using video captions,'' \emph{arXiv
  preprint arXiv:2107.00941}, 2021.

\bibitem{wang2022veracity}
S.~Wang, X.~Xu, X.~Zhang, Y.~Wang, and W.~Song, ``Veracity-aware and
  event-driven personalized news recommendation for fake news mitigation,'' in
  \emph{Proceedings of the ACM Web Conference 2022}, 2022, pp. 3673--3684.

\bibitem{you2019attributed}
D.~You, V.~Nguyen, K.~Lee, and Q.~Liu, ``Attributed multi-relational attention
  network for fact-checking url recommendation,'' in \emph{Proceedings of the
  28th ACM International Conference on Information and Knowledge Management},
  2019, pp. 1471--1480.

\bibitem{kempe2003maximizing}
D.~Kempe, J.~Kleinberg, and {\'E}.~Tardos, ``Maximizing the spread of influence
  through a social network,'' in \emph{Proceedings of the ninth ACM SIGKDD
  international conference on Knowledge discovery and data mining}, 2003, pp.
  137--146.

\bibitem{nguyen2012containment}
N.~Nguyen, G.~Yan, M.~Thai, and S.~Eidenbenz, ``Containment of misinformation
  spread in online social networks,'' in \emph{Proceedings of the 4th Annual
  ACM Web Science Conference}, 2012, pp. 213--222.

\bibitem{vo2018rise}
V.~Nguyen and K.~Lee, ``The rise of guardians: Fact-checking url recommendation
  to combat fake news,'' in \emph{The 41st international ACM SIGIR conference
  on research \& development in information retrieval}, 2018, pp. 275--284.

\bibitem{hossain2020covidlies}
T.~Hossain, R.~L. Logan~IV, A.~Ugarte, Y.~Matsubara, S.~Young, and S.~Singh,
  ``Covidlies: Detecting covid-19 misinformation on social media,'' in
  \emph{Workshop on NLP for COVID-19 (Part 2) at EMNLP 2020}, 2020.

\bibitem{alhakami2022evaluating}
H.~Alhakami, W.~Alhakami, A.~Baz, M.~Faizan, M.~W. Khan, and A.~Agrawal,
  ``Evaluating intelligent methods for detecting covid-19 fake news on social
  media platforms,'' \emph{Electronics}, vol.~11, no.~15, p. 2417, 2022.

\bibitem{kou2022hc}
Z.~Kou, L.~Shang, Y.~Zhang, and D.~Wang, ``Hc-covid: A hierarchical crowdsource
  knowledge graph approach to explainable covid-19 misinformation detection,''
  \emph{Proceedings of the ACM on Human-Computer Interaction}, vol.~6, no.
  GROUP, pp. 1--25, 2022.

\bibitem{verma2022examining}
G.~Verma, A.~Bhardwaj, T.~Aledavood, M.~De~Choudhury, and S.~Kumar, ``Examining
  the impact of sharing covid-19 misinformation online on mental health,''
  \emph{Scientific Reports}, vol.~12, no.~1, p. 8045, 2022.

\bibitem{germani2022and}
F.~Germani, A.~B. Pattison, and M.~Reinfelde, ``Who and digital agencies: how
  to effectively tackle covid-19 misinformation online,'' \emph{BMJ Global
  Health}, vol.~7, no.~8, p. e009483, 2022.

\bibitem{pennington2014glove}
J.~Pennington, R.~Socher, and C.~Manning, ``Glove: Global vectors for word
  representation,'' in \emph{Proceedings of the 2014 conference on empirical
  methods in natural language processing (EMNLP)}, 2014, pp. 1532--1543.

\end{thebibliography}
